\documentclass[runningheads]{llncs}
\usepackage{graphicx}
\usepackage[shortlabels]{enumitem}
\usepackage{booktabs}
\usepackage{threeparttable}
\usepackage{multirow}
\usepackage{amsfonts}
\usepackage[table]{xcolor}
\usepackage{amsmath,amsfonts,bm}
\usepackage{algorithm}
\usepackage{algorithmic}
\usepackage{color}
\usepackage{comment}
\usepackage{pifont}
\usepackage{tikz}
\usepackage{hyperref}
\newcommand{\cmark}{\text{\ding{51}}}
\newcommand{\xmark}{\text{\ding{55}}}
\newcommand{\minitab}[2][l]{\begin{tabular}{#1}#2\end{tabular}} 
\def\code#1{\texttt{#1}}
\newcommand{\tikzcircle}[2][red,fill=red]{\tikz[baseline=-0.5ex]\draw[#1,radius=#2] (0,0) circle ;}%
\hypersetup{
    colorlinks=true,
    linkcolor=blue,
    filecolor=magenta,      
    urlcolor=cyan,
    citecolor=teal
}
\begin{document}
\title{Simple, Scalable, and Stable Variational Deep Clustering
%\thanks{This work is supported by ...}
}

\author{Lele Cao\inst{1} \and
Sahar Asadi\inst{1} \and
Wenfei Zhu\inst{1,2} \and
Christian Schmidli\inst{1} \and
Michael Sj\"{o}berg\inst{1}}
\authorrunning{Cao L., Asadi S., et al.}
% First names are abbreviated in the running head.
% If there are more than two authors, 'et al.' is used.
%
\institute{King Digital Entertainment, Activision Blizzard Group, Stockholm, Sweden
\email{\{lele.cao,sahar.asadi,wenfei.zhu,christian.schmidli,michael.sjoberg\}@king.com} \and
%\email{\{firstname.lastname\}@king.com} \and
Department of Information Technology, Uppsala University, Uppsala, Sweden}

\maketitle % typeset the header of the contribution
\begin{abstract}
Deep clustering (DC) has become the state-of-the-art for unsupervised clustering. In principle, DC represents a variety of unsupervised methods that jointly learn the underlying clusters and the latent representation directly from unstructured datasets. However, DC methods are generally poorly applied due to high operational costs, low scalability, and unstable results. In this paper, we first evaluate several popular DC variants in the context of industrial applicability using eight empirical criteria. We then choose to focus on variational deep clustering (VDC) methods, since they mostly meet those criteria except for simplicity, scalability, and stability. To address these three unmet criteria, we introduce four generic algorithmic improvements: initial $\gamma$-training, periodic $\beta$-annealing, mini-batch GMM (Gaussian mixture model) initialization, and inverse min-max transform. We also propose a novel clustering algorithm S3VDC (simple, scalable, and stable VDC)\footnote{Source code: \url{https://github.com/king/s3vdc}} 
that incorporates all those improvements. Our experiments show that S3VDC outperforms the state-of-the-art on both benchmark tasks and a large unstructured industrial dataset without any ground truth label. In addition, we analytically evaluate the usability and interpretability of S3VDC.

\end{abstract}

%% === SEC: INTRODUCTION ==============
\section{Introduction}
\label{intro}
Clustering algorithms aim to group a set of data points into clusters such that: 1) points within each cluster are similar, and 2) points from different clusters are dissimilar. Traditional clustering methods like $k$-means and Gaussian mixture models (GMM) are, however, highly dependent on the input data; hence, they are ineffective when the input dimensionality is very high. Formerly, feature extraction methods such as dimension reduction and representation learning were extensively applied as a step prior to clustering.
Recent works show that optimizing clustering jointly with feature extraction using deep neural networks (DNNs) yields superior results~\cite{guo2017deep,jiang2017variational,mukherjee2019clustergan}. This paradigm is usually termed deep clustering (DC). 
Despite the high evaluation scores achieved by DC approaches reported on a few benchmark datasets (Table~\ref{tab:evalDC}), DC has not shown consistent success on large dynamic industrial datasets. We empirically discovered that the following properties are crucial to the \textit{industrial applicability} of DCs. This motivates the active adoption of unsupervised DC algorithms for exploring large-scale industrial datasets.
\begin{enumerate}[P1.]
\item Truly unsupervised: unavailability of ground-truth label should not significantly compromise the credibility of model performance. 
\item Access to latent embedding: this is typically important to understand the feature space. In addition, the learned embedding can be utilized as input to other machine learning (ML) applications.
\item Indicator of the optimal number of clusters: 
there should be a natural way to determine the best number of clusters without any label guidance.
\item Generative: the model allows generating samples from the underlying data distribution to enable tasks such as data augmentation and cluster interpretation.
\item Learnable cluster weights: the optimal cluster weights (the relative likelihood of each cluster) should be learned without enforcing any constant prior. 
\item \textit{Simplicity}: once implemented (in a preferably end-to-end manner, and independent of any pretrained model or sequential steps) and verified, the manual operational cost stays constantly and continuously at a low level.
\item \textit{Scalability}: the computational complexity and memory efficiency remain largely invariant to both the size of datasets and the number of clusters.
\item \textit{Stability}: favorable results with low variance should be guaranteed from different trials with random initialization so that the frequent model iteration process (subsequent training on newly accumulated data) becomes easier and more efficient.
\end{enumerate}
We examine qualitatively the state-of-the-art DC methods on P1-P5 in Section~\ref{evalDC} and show that variational deep clustering (VDC) meets these properties the best. Section~\ref{briefVDC} presents a unified introduction to the mainstream VDC algorithms. In Section~\ref{S3VDC}, we discuss four common problems of VDCs that violate P6-P8, and we propose generic solutions to address each of those problems. Finally, Section~\ref{sec:exp} presents a comprehensive experimental and exploratory analysis. The main contributions of this work are:

\begin{itemize}
\item We identify essential properties for the industrial applicability of DC. In addition, we present a qualitative evaluation of state-of-the-art DC algorithms and recommend VDC model family as the most suitable approach.
\item We propose practical and generic solutions to address the \textit{simplicity}, \textit{scalability}, and \textit{stability} problems in VDC baselines, resulting in an industrial-friendly VDC algorithm: S3VDC (simple, scalable, and stable VDC).
\item Our experimental results show that S3VDC outperforms baseline VDCs on scalability and stability. Furthermore, our exploratory analysis demonstrates how other industrial ``nice-to-have'' properties can be fulfilled.
\end{itemize}

\begin{table}[tb!]
\footnotesize\addtolength{\tabcolsep}{2pt}\renewcommand{\arraystretch}{1.1}
\begin{center}
\caption{
Qualitative evaluation of DC algorithms with respect to properties P1-P8 (Section~\ref{intro}). ``\cmark''~means that the method has the property by design. ``\xmark''~denotes that the model prohibits the property. ``\textbf{H}'' (high) and ``\textbf{L}'' (low) denote the required effort to satisfy the property. ``\textbf{-}'' indicates cases where we could not conclude with reasonable effort. NMI (normalized mutual information) is a supervised clustering metric.
}
\label{tab:evalDC}
\begin{threeparttable}
\begin{tabular}{c|c|c|c|c|c|c|c|c|c|c|c}
\bottomrule
\multicolumn{2}{c|}{\multirow{2}{*}{\minitab[c]{DC methods vs.\\preferred properties}}}
 & \multirow{2}{*}{P1} 
 & \multirow{2}{*}{P2}
 & \multirow{2}{*}{P3}
 & \multirow{2}{*}{P4}
 & \multirow{2}{*}{P5}
 & \multirow{2}{*}{P6}
 & \multirow{2}{*}{P7}
 & \multirow{2}{*}{P8} 
 & \multicolumn{2}{c}{MNIST Dataset}
% & \minitab[c]{1. truly\\unsupervised} 
% & \minitab[c]{2. latent\\embedding}  
% & \minitab[c]{3. optimal\\cluster num.}  
% & \minitab[c]{4. generative\\capability}  
% & \minitab[c]{5. cluster\\weights}
% & 6. simplicity
% & 7. scalability
% & 8. stability 
\\
\cline{11-12}
\multicolumn{2}{c|}{}&&&&&&&&& Accuracy & NMI 
\\
\hline
\multirow{5}{*}{\minitab[c]{DNN\\based\\DC}} 
& DEC~\cite{xie2016unsupervised} & \cellcolor{green!30!black}\textcolor{white}\cmark & \cellcolor{green!30!black}\textcolor{white}\cmark & \cellcolor{red!15}H & \cellcolor{black!5}\xmark & \cellcolor{red!15}H & \cellcolor{blue!35}L & \cellcolor{red!15}H & \cellcolor{red!15}H\tnote{*} & 0.84 & 0.80\\
\cline{2-12}
& DCEC~\cite{guo2017deep} & \cellcolor{green!30!black}\textcolor{white}\cmark & \cellcolor{green!30!black}\textcolor{white}\cmark & \cellcolor{red!15}H & \cellcolor{black!5}\xmark & \cellcolor{red!15}H & \cellcolor{blue!35}L & \cellcolor{red!15}H & \cellcolor{blue!35}L\tnote{*} & 0.88 & 0.88 \\
\cline{2-12}
& JULE~\cite{yang2016joint} & \cellcolor{green!30!black}\textcolor{white}\cmark & \cellcolor{green!30!black}\textcolor{white}\cmark & \cellcolor{red!15}H & \cellcolor{black!5}\xmark & \cellcolor{red!15}H & \cellcolor{red!15}H & \cellcolor{red!15}H & - & - & 0.91 \\
\cline{2-12}
& DEPICT~\cite{ghasedi2017deep} & \cellcolor{green!30!black}\textcolor{white}\cmark & \cellcolor{green!30!black}\textcolor{white}\cmark & \cellcolor{red!15}H & \cellcolor{black!5}\xmark & \cellcolor{red!15}H & \cellcolor{red!15}H & \cellcolor{red!15}H & \cellcolor{red!15}H & 0.96 & 0.91 \\
\cline{2-12}
& IMSAT~\cite{hu2017learning} & \cellcolor{blue!35}L & \cellcolor{blue!35}L & \cellcolor{red!15}H & \cellcolor{black!5}\xmark & \cellcolor{red!15}H & \cellcolor{blue!35}L & \cellcolor{blue!35}L & \cellcolor{red!15}H & 0.98 & - \\
\hline
\multirow{4}{*}{\minitab[c]{GAN\\based\\DC}} 
& InfoGAN~\cite{chen2016infogan} & \cellcolor{blue!35}L & \cellcolor{blue!35}L & \cellcolor{red!15}H & \cellcolor{green!30!black}\textcolor{white}\cmark & \cellcolor{red!15}H & \cellcolor{blue!35}L & \cellcolor{green!30!black}\textcolor{white}\cmark & \cellcolor{red!15}H & 0.89~\cite{mukherjee2019clustergan} & 0.86~\cite{mukherjee2019clustergan} \\
\cline{2-12}
& Sub-GAN~\cite{liang2018sub} & \cellcolor{green!30!black}\textcolor{white}\cmark & \cellcolor{green!30!black}\textcolor{white}\cmark & \cellcolor{red!15}H & \cellcolor{green!30!black}\textcolor{white}\cmark & \cellcolor{red!15}H & \cellcolor{blue!35}L & \cellcolor{red!15}H & \cellcolor{red!15}H & 0.85 & - \\
\cline{2-12}
& ClusterGAN~\cite{mukherjee2019clustergan} & \cellcolor{green!30!black}\textcolor{white}\cmark & \cellcolor{blue!35}L & \cellcolor{red!15}H & \cellcolor{green!30!black}\textcolor{white}\cmark & \cellcolor{red!15}H & \cellcolor{blue!35}L & \cellcolor{green!30!black}\textcolor{white}\cmark & \cellcolor{red!15}H & 0.95 & 0.89 \\
\cline{2-12}
& ClusterGAN~\cite{ghasedi2019balanced} & \cellcolor{green!30!black}\textcolor{white}\cmark & \cellcolor{blue!35}L & \cellcolor{red!15}H & \cellcolor{green!30!black}\textcolor{white}\cmark & \cellcolor{red!15}H & \cellcolor{red!15}H & \cellcolor{red!15}H & - & 0.96 & 0.92 \\
\hline
\multirow{5}{*}{\minitab[c]{VAE\\based\\DC\\\scriptsize(VDC)}} 
& M1+M2~\cite{kingma2014semi}\tnote{$\dagger$} & \cellcolor{blue!35}L & \cellcolor{blue!35}L & \cellcolor{blue!35}L & \cellcolor{green!30!black}\textcolor{white}\cmark & \cellcolor{red!15}H & \cellcolor{blue!35}L & \cellcolor{green!30!black}\textcolor{white}\cmark & \cellcolor{red!15}H & - & - \\
\cline{2-12}
& M1+2~\cite{figueroasimple}\tnote{$\ddagger$} & \cellcolor{green!30!black}\textcolor{white}\cmark & \cellcolor{green!30!black}\textcolor{white}\cmark & \cellcolor{blue!35}L & \cellcolor{green!30!black}\textcolor{white}\cmark & \cellcolor{red!15}H & \cellcolor{blue!35}L & \cellcolor{blue!35}L & \cellcolor{red!15}H & 0.83 & 0.80 \\
\cline{2-12}
& DLGMM~\cite{nalisnick2016approximate}\tnote{$\dagger$} & \cellcolor{blue!35}L & \cellcolor{green!30!black}\textcolor{white}\cmark & \cellcolor{blue!35}L & \cellcolor{green!30!black}\textcolor{white}\cmark & \cellcolor{red!15}H & \cellcolor{red!15}H & \cellcolor{blue!35}L & \cellcolor{red!15}H & - & - \\
\cline{2-12}
& GMVAE~\cite{dilokthanakul2016deep} & \cellcolor{green!30!black}\textcolor{white}\cmark & \cellcolor{green!30!black}\textcolor{white}\cmark & \cellcolor{blue!35}L & \cellcolor{green!30!black}\textcolor{white}\cmark & \cellcolor{red!15}H & \cellcolor{red!15}H & \cellcolor{red!15}H & \cellcolor{red!15}H & 0.82 & - \\
\cline{2-12}
& VaDE~\cite{jiang2017variational} & \cellcolor{green!30!black}\textcolor{white}\cmark & \cellcolor{green!30!black}\textcolor{white}\cmark & \cellcolor{blue!35}L & \cellcolor{green!30!black}\textcolor{white}\cmark & \cellcolor{green!30!black}\textcolor{white}\cmark & \cellcolor{red!15}H\tnote{*} & \cellcolor{red!15}H\tnote{*} & \cellcolor{red!15}H\tnote{*} & 0.94 & - \\
\toprule
\end{tabular}
\begin{tablenotes}
        \scriptsize
        \item[*] Evaluated using the open-source implementation of the respective method.
        \item[$\dagger$] Strictly, these methods can not be regarded as unsupervised DC approaches.
        \item[$\ddagger$] We use ``M1+2'' to refer to the simplified and unsupervised VDC~\cite{figueroasimple} adapted from the original M1+M2~\cite{kingma2014semi}.
\end{tablenotes}
\end{threeparttable}\end{center}
\end{table}

%% === SEC: REVIEW & EVALUATION of DC ==============
\section{Review and Evaluation of Deep Clustering (DC)}
\label{evalDC}
DC models fall into three primary categories according to their network architecture: DNN, generative adversarial network (GAN), and variational auto-encoder (VAE) based DC, among which GAN and VAE are deep generative approaches; thus, they can capture high dimensional probability distributions, impute missing data, and deal with multi-modal outputs. Table~\ref{tab:evalDC} presents a qualitative evaluation of the state-of-the-art DC methods with respect to properties P1-P8. 

The DNN-based category includes DC approaches that apply multi-layer perceptron (MLP), deep belief network (DBN), or convolutional neural network (CNN). 
Deep embedded clustering (DEC)~\cite{xie2016unsupervised} is the first well-known DC method. It learns the representations sequentially using stacked auto-encoder (AE), initializes clusters with $k$-means, and finetunes the encoder with a clustering loss.~\cite{guo2017deep} proposed DCEC, which improves DEC by preserving its decoder and employing CNN as a feature extractor. 
Among other influential methods in this category are JULE~\cite{yang2016joint}, DEPICT~\cite{ghasedi2017deep}, and IMSAT~\cite{hu2017learning}. Lack of P4 (generating new samples) and P5 (modeling cluster weights) is the primary issue with DNN-based DCs due to their underlying architecture (see Table~\ref{tab:evalDC}). 

GAN-based methods contain a system of two DNNs (generator and discriminator) that play a zero-sum game~\cite{goodfellow2014generative}. 
InfoGAN~\cite{chen2016infogan}, a highly cited GAN-based DC baseline, optimizes the mutual information of latent variables constructed by a mixture of Gaussian (MoG). 
In~\cite{liang2018sub}, a novel clustering model, Sub-GAN, was prposed. Recently, two GAN-based DC approaches~\cite{mukherjee2019clustergan,ghasedi2019balanced}, both named after ClusterGAN, were proposed:~\cite{mukherjee2019clustergan} considers discrete-continuous mixtures to sample noise variable, while~\cite{ghasedi2019balanced} utilizes conditional entropy minimization as clustering loss combined with a few tricks to enable self-paced and balanced training. 
GAN-based methods may suffer from catastrophic model collapse as well as unstable training (P8). Moreover, they require additional discriminator networks (P6). They also struggle to determine the optimal number of clusters and their weights (P3, P5).

VAE is a generative variant of AE that forces latent variables to follow a predefined distribution. 
In Table~\ref{tab:evalDC}, M1+M2~\cite{kingma2014semi} and DLGMM~\cite{nalisnick2016approximate} are two mixture VAEs, which may be applied to unsupervised clustering tasks with modest adaptations. M1+M2 tackles semi-supervised classification problems using hierarchical stochastic layers that are fundamentally cumbersome to train. DLGMM uses MoG to approximate VAE posterior but does not model the cluster variable. Recent work directly combines VAE with GMMs, producing a few well-known VAE-based DC (a.k.a.~VDC) approaches: GMVAE~\cite{dilokthanakul2016deep}, M1+2~\cite{figueroasimple}, and VaDE~\cite{jiang2017variational}. VDC methods satisfy more industrial properties compared to DNN and GAN based ones except for simplicity, scalability, and stability (P6-P8). The next sections focus on making VDC approaches comply with P6-P8.

%% === SEC: VDC ALGORITHMS ==============
\section{Variational Deep Clustering (VDC) Algorithms}
\label{briefVDC}

VDCs try to generatively model an unlabeled dataset $\mathbf{X}=\{\mathbf{x}^{(n)}\}_{n=1}^N$ under the constraint of $C$ clusters; $N$ denotes the total number of samples in $\mathbf{X}$. For the sake of conciseness, we use the general term $\mathbf{x}\in\mathbb{R}^D$ to denote any $D$-dimensional data sample when we walk-through the VDC methods mentioned in Section~\ref{evalDC}: M1+2~\cite{figueroasimple}, GMVAE~\cite{dilokthanakul2016deep}, and VaDE~\cite{jiang2017variational}.

%----------
\subsection{Generative Processes}
The generative step of VDCs starts with sampling a cluster $c$ from a categorical distribution $p(c)$ parameterized by $\bm{\pi}\!\in\!\mathbb{R}^C_+$:
\begin{equation}\label{eq:sampleC}
c\sim Cat(\bm{\pi}), \text{~~~~~~s.t.:} \sideset{}{_{c=1}^C}\sum\pi_c=1\ ,
\end{equation} 
where $\pi_c$ is the prior probability (i.e. cluster weights) for the $c$-th cluster. Note that VaDE treats $\pi_c$ as trainable GMM parameters while others set them uniformly to $C^{-1}$. Next, a latent vector $\mathbf{z}$ is chosen from $p(\mathbf{z})$ via one of% the following processes:
\begin{flalign}
\text{M1+2: }&\:\mathbf{z}\sim \mathcal{N}(0, I)\ , \\
\text{GMVAE: }&\:\mathbf{z}'\sim \mathcal{N}(0, \mathbf{I}), \mathbf{z}\sim \mathcal{N}(\bm{\mu}_c(\mathbf{z}'), \bm{\sigma}_c^2(\mathbf{z}')) \ , \\
\text{VaDE: }&\:\mathbf{z}\sim \mathcal{N}(\bm{\mu}_c, \bm{\sigma}_c^2\mathbf{I})\ ,
\end{flalign}
where M1+2 uses a single multivariate Gaussian $\mathcal{N}(\cdot)$, which may lower the upper bound of its performance; $\mathbf{I}$ is an identity matrix; $\bm{\mu}_c$ and $\bm{\sigma}_c^2$ are the mean and variance of the Gaussian for the $c$-th cluster, respectively. VaDE initializes those as global parameters with a GMM. In GMVAE, if we marginalize out $\mathbf{z}'$, $\mathbf{z}$ is an arbitrary distribution parameterized by functions $\bm{\mu}_c(\cdot)$ and $\bm{\sigma}^2_c(\cdot)$ approximated by neural networks (NNs). Thus, it does not scale well to large $C$ values due to the larger network size required to represent each Gaussian.

VDCs then compute $\bm{\mu}_x$ (and also $\bm{\sigma}_x^2$ if $\mathbf{x}$ is real-valued) via functions approximated by density NNs parameterized by $\bm{\theta}$, i.e. $f_{\bm{\theta}}(\mathbf{z};c)$ for M1+2 or $f_{\bm{\theta}}(\mathbf{z})$ otherwise. Finally, a sample $\mathbf{x}$ is selected from:
\begin{equation}\label{eq:sampleX}
\mathbf{x}\sim \mathcal{N}(\bm{\mu}_x, \bm{\sigma}_x^2\mathbf{I}) \text{~~~~or~~~~} \mathcal{B}(\bm{\mu}_x),
\end{equation} 
where $\mathcal{B}(\cdot)$ is a Bernoulli distribution parameterized by $\bm{\mu}_x$ when $\mathbf{x}$ is binary. For simplicity, we proceed with the assumption that sample $\mathbf{x}$ is binary. 

%----------
\subsection{Variational Lower Bound Objectives}
VDCs are usually optimized through DNNs, so the stochastic gradient variational Bayes (SGVB) estimator and the reparameterization trick~\cite{diederik2014auto} can be used to maximize the log-evidence lower bound (ELBO). As shown previously, M1+2 considers the generative model $p(\mathbf{x},\mathbf{z},c)=p(\mathbf{x}|\mathbf{z},c)p(\mathbf{z})p(c)$, and the posterior $p(\mathbf{z},c|\mathbf{x})$ is approximated with a tractable mean-field distribution $q(\mathbf{z},c|\mathbf{x})=q(\mathbf{z}|\mathbf{x})q(c|\mathbf{x})$. So, its ELBO is
\begin{equation}
\label{eq:ELBO_M12}
\begin{aligned}
\mathcal{L}_{\text{M1+2}}\!=&\mathbb{E}_{q(\mathbf{z},c|\mathbf{x})}[\ln (p(\mathbf{x},\mathbf{z},c)/q(\mathbf{z},c|\mathbf{x}))] \\
\!=&\mathbb{E}_{q(\mathbf{z},c|\mathbf{x})}[\ln p(\mathbf{x}|\mathbf{z},c)]\!-\mathcal{D}(q(c|\mathbf{x})||p(c))-\mathcal{D}(q(\mathbf{z}|\mathbf{x})||p(\mathbf{z})), 
\end{aligned}	
\end{equation} 
where the first term is the negative reconstruction loss, and the terms followed are Kullback-Leibler (KL) divergence functions, noted as $\mathcal{D}(\cdot||\cdot)$, which regularize the categorical and Gaussian distributions. Similarly, VaDE uses the generative model $p(\mathbf{x},\mathbf{z},c)=p(\mathbf{x}|\mathbf{z})p(\mathbf{z}|c)p(c)$, hence the ELBO is
\begin{equation}
\label{eq:ELBO_VaDE}
\begin{aligned}
\mathcal{L}_{\text{VaDE}}\!=&\mathbb{E}_{q(\mathbf{z},c|\mathbf{x})}[\ln p(\mathbf{x}|\mathbf{z})]-\mathcal{D}(q(\mathbf{z},c|\mathbf{x})||p(\mathbf{z},c)) \\
\!=&\mathbb{E}_{q(\mathbf{z},c|\mathbf{x})}[\ln p(\mathbf{x}|\mathbf{z})]-\mathcal{D}(q(c|\mathbf{x})||p(c))-\mathcal{D}(q(\mathbf{z}|\mathbf{x})||p(\mathbf{z}|c)), 
\end{aligned}	
\end{equation}
where the first part is the reconstruction term and the rest regularize the latent posterior $q(\mathbf{z},c|\mathbf{x})$ to lie on a manifold of MoG prior $p(\mathbf{z},c)$. GMVAE considers the generative model $p(\mathbf{x},\mathbf{z},\mathbf{z}',c)=p(\mathbf{x}|\mathbf{z})p(\mathbf{z}|\mathbf{z}',c)p(\mathbf{z}')p(c)$ and a mean-field posterior proxy $q(\mathbf{z},\mathbf{z}',c|\mathbf{x})=q(\mathbf{z}|\mathbf{x})q(\mathbf{z}'|\mathbf{x})p(c|\mathbf{z},\mathbf{z}')$, thus the ELBO can be written as
\begin{equation}
\label{eq:ELBO_GMVAE}
\begin{aligned}
\mathcal{L}_{\text{GMVAE}}\!=&\mathbb{E}_{q(\mathbf{z},\mathbf{z}',c|\mathbf{x})}[\ln (p(\mathbf{x},\mathbf{z},\mathbf{z}',c)/q(\mathbf{z},\mathbf{z}',c|\mathbf{x}))] \\
\!=&\mathbb{E}_{q(\mathbf{z},\mathbf{z}',c|\mathbf{x})}[\ln p(\mathbf{x}|\mathbf{z})]\\
&-\!\mathcal{D}(p(c|\mathbf{z},\mathbf{z}')||p(c))\!-\!\mathcal{D}(q(\mathbf{z}'|\mathbf{x})||p(\mathbf{z}'))\!-\!\mathcal{D}(q(\mathbf{z}|\mathbf{x})||p(\mathbf{z}|\mathbf{z}',c))\ , \\
\end{aligned}
\end{equation} 
where the three terms denote the reconstruction error and regularizers of $c$-prior, $\mathbf{z}'$-prior, and conditional prior, respectively. Universally in ELBO objectives of VDC methods, $p(\mathbf{x}|\mathbf{z})$, $q(\mathbf{z}|\mathbf{x})$, and $q(\mathbf{z}'|\mathbf{x})$ are respectively modelled with DNNs approximating functions of $f_{\bm{\theta}}(\mathbf{z})$, $g_{\bm{\phi}}(\mathbf{x})$, and $g_{\bm{\phi}'}(\mathbf{x})$, where $\bm{\theta}$, $\bm{\phi}$, and $\bm{\phi}'$ are their trainable parameters. 

%% === SEC: S3VDC ==============
\section{Simple, Scalable, and Stable VDC}
\label{S3VDC}
In this section, we discuss four frequently encountered problems that violate the \textit{simplicity}, \textit{scalability}, and \textit{stability} properties (P6-P8 in Section~\ref{intro}), and propose generic solutions with pertinence, which lead to a holistic algorithm S3VDC.

\subsection{Initial $\mathbf{\gamma}$-Training: Reproduce with Milder Volatility}
\label{sec:gammatraining}
The necessity of VDC reproducibility is beyond doubt. However, we found that the clustering results of VDCs are heavily affected by randomness (e.g., in parameter initialization and data pipeline implementation), which agrees with the pattern reported by~\cite{47692}. The evaluation of VDCs towards the stability property (P8) in Table~\ref{tab:evalDC} shows that the objective functions of VDC have little impact on stability. A simple remedy is to pretrain networks $f_{\bm{\theta}}(\mathbf{z})$, $g_{\mathbf{\phi}}(\mathbf{x})$, and $g_{\mathbf{\phi}'}(\mathbf{x})$ using a stacked AE that has the same network architecture and reconstruction loss~\cite{jiang2017variational}. Once pretrained, the weights are copied to the original VDC model before finetuning the fully fleshed ELBO objectives. Unfortunately, this practice improves the stability at the cost of simplicity due to requiring 1) multiple sequential steps, 2) higher space complexity, and 3) maintaining multiple pretrained models for different datasets.

\begin{figure}
\centering
\scalebox{0.75}{\includegraphics{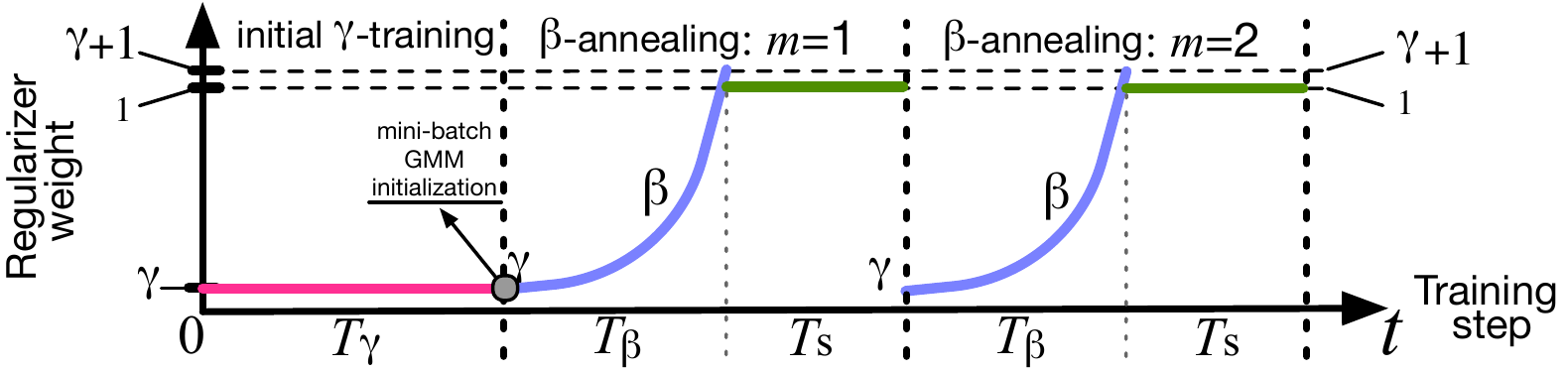}}
\caption{An overview of the training process in S3VDC algorithm: initial $\mathbf{\gamma}$-training, mini-batch GMM initialization, and periodic $\mathbf{\beta}$-annealing.}
\label{fig:reg_anneal}
\end{figure}

The na\"ive ELBO optimization targets for VDCs consist of two parts (see Equations~(\ref{eq:ELBO_M12}-\ref{eq:ELBO_GMVAE})): 1) a reconstruction term that guarantees data reconstruction capability of the latent representation $\mathbf{z}$, and 2) several KL regularization terms to leverage the prior knowledge. Authors of~\cite{chou2019generated} concluded that higher impact of regularizers pushes VAE to posterior collapse, while less regularization makes VAE more interchangeable with a plain AE. Consequently, we propose an initial ``$\gamma$-training'' phase, indicated by the straight red line in Figure~\ref{fig:reg_anneal}, that trains VDCs with a much lower emphasis (weighted by $0<\!\gamma\!\ll\!1$) on the regularizer terms for the first $T_\gamma$ mini-batch steps:
\begin{equation}
\label{eq:gammaPretrain}
\mathcal{L}_\gamma=\mathbb{E}_{q(\cdot|\hat{\mathbf{x}})}[\ln p(\mathbf{x}|\cdot)]-\gamma[\mathcal{D}(q(\cdot|\hat{\mathbf{x}})||\cdot)+\ldots],\ T_\gamma\geq t>0\ ,
\end{equation} 
where $t$ is the index of the current training step. We also apply the denoising mechanism, in which $\mathbf{x}$ is corrupted into $\hat{\mathbf{x}}\!\in\!\hat{\mathbf{X}}$ by adding noise from a zero-mean random normal distribution. In this way, VDCs reconstruct a ``repaired'' input from a corrupted version. With this, pretraining is fused into the VDC optimization processes, which brings not only \textit{stable} results but also \textit{simple} maintenance.

%----------
\subsection{Periodic $\mathbf{\beta}$-Annealing: Improve Latent Disentanglement}
We empirically observed that VDCs are vulnerable to model degeneration where the optimization focuses too much on KL regularizers from a certain point and never escapes, leading to uninformative $\mathbf{z}$ that jeopardizes the learning of disentangled latent representations. The intuitive solution is to carefully schedule the regularization weight (a.k.a. regularizer annealing). Similar modification is made to VAEs~\cite{bowman2016generating,higgins2017beta,liu2019cyclical}. The key to designing the annealing policy lies between the intricate balance of two seemingly contradictory factors: 1) the provision of meaningful $\mathbf{z}$ before training $f_{\bm{\theta}}(\mathbf{z})$, and 2) continuously leveraging better $\mathbf{z}$ distribution. To that end, we propose a periodic polynomial annealing strategy, ``periodic $\beta$-annealing'', after optimizing Equation~\eqref{eq:gammaPretrain} for $T_\gamma$ steps. As illustrated in Figure~\ref{fig:reg_anneal}, any period $m$ has two phases: 1) $\beta$-annealing for $T_\beta$ steps optimizing the following objective, and 2) training with na\"ive VDC objectives for $T_{\text{s}}$ steps.
\begin{equation}
\label{eq:anneal}
\begin{aligned}
\mathcal{L}_\beta\!\!=&\mathbb{E}_{q(\cdot|\hat{\mathbf{x}})}[\ln p(\mathbf{x}|\cdot)]-\beta[\mathcal{D}(q(\cdot|\hat{\mathbf{x}})||\cdot)+\ldots] \ ,
\\
\text{s.t.: }&\beta=\gamma+\left[\frac{t-T_\gamma-(m-1)(T_\beta+T_{\text{s}})}{T_\beta}\right]^u \ ,\\
&T_\gamma+(m-1)(T_\beta+T_{\text{s}}) + T_\beta \geq t>T_\gamma+(m-1)(T_\beta+T_{\text{s}}) \ .
\end{aligned}	
\end{equation} 
$\beta$ represents the scheduled weight of KL regularizers during the first phase in each period; $m\in\{1\ldots M\}$ denotes the $m$-th annealing period. To \textit{stabilize} the training as much as possible, $u$ should be selected to ensure both a smooth transition from $\gamma$-training and a mild increase of $\beta$. We empirically find that periodic $\beta$-annealing improves latent disentanglement in practice, therefore, it further improves the clustering \textit{stability}. 

%----------
\subsection{Mini-Batch GMM Initialization: Scale to Large Datasets}
\label{proposalScale}
For VDCs to work on large-scale datasets, their performance should be largely invariant to both the total number of samples $N$ and the number of clusters $C$. M1+2 meets both conditions at the expense of restricting to using a single Gaussian. GMVAE uses one NN to approximate a tuple ($\bm{\mu}_c$, $\bm{\sigma}^2_c$), so its complexity increases linearly with $C$. As a result, M1+2 and GMVAE have much inferior performance compared to VaDE (see accuracy and NMI scores in Table~\ref{tab:evalDC}).
On the other hand, the computational bottleneck of VaDE lies in the step where $\pi_c$, $\bm{\mu}_c$ and $\bm{\sigma}^2_c$ are initialized using all $N$ samples. To improve the \textit{scalability} of VaDE, we merely take $k\!\times\!L$ ($\ll\!N$) Monte Carlo samples from the entire dataset to initialize GMM (represented with an anchor ``\tikzcircle[black, fill=gray]{2.5pt}''in Figure~\ref{fig:reg_anneal}); $L$ denote the mini-batch size, hence the name ``mini-batch GMM initialization''. Section~\ref{exp:scalability} shows how robust this simplification is in relation to the value of $k\in\{1,2,\ldots\}$. As an auxiliary study, we also tried to update $\pi_c$, $\bm{\mu}_c$ and $\bm{\sigma}^2_c$ (using $k\!\times\!L$ samples for every GMM initialization) before each $\beta$-annealing period with a momentum factor (similar to~\cite{cao2017real}), yet it did not offer observable uplift of clustering accuracy.

%----------
\subsection{Inverse Min-Max Transform: Avoid \code{NaN} Losses}
\label{sec:minmaxNorm}
\code{NaN} loss values are not uncommon when training VAE and VDC models using ML frameworks like Tensorflow and PyTorch. Spinning up a large number of parallel training sessions to guarantee sufficient successful ones is a waste of computing resources. Worst of all, there is no easy way to escape from \code{NaN} loss if pretraining is considered as a standalone step, which is the situation we end up with when experimenting with certain VDC models such as~\cite{jiang2017variational}. 

Long story short, the curse of \code{NaN} losses originates from the terms of $q(c|\mathbf{x})$ and $q(c|\mathbf{z},\mathbf{z}')$ in Equations~\eqref{eq:ELBO_M12}, \eqref{eq:ELBO_VaDE}, and \eqref{eq:ELBO_GMVAE}. The authors of~\cite{jiang2017variational} illustrate that $q(c|\mathbf{x})$ can be approximated with $q(c|\mathbf{z})$ in VaDE:
\begin{equation}\label{eq:qcxApprox}
	q(c|\mathbf{x}) \approx p(c|\mathbf{z}) \equiv \frac{p(c)e^{\ln p(\mathbf{z}|c)}}{\sum_{c'=1}^C p(c')e^{\ln p(\mathbf{z}|c')}}\ ,
\end{equation}
where $p(\mathbf{z}|c)=\mathcal{N}(\mathbf{z}|\bm{\mu}_c, \bm{\sigma}_c^2\mathbf{I})$. For the sake of numerical stability, we formulate $p(\mathbf{z}|c)$ as $e^{\ln p(\mathbf{z}|c)}$ instead. The log probability $\ln p(\mathbf{z}|c)$ is directly calculated with
\begin{equation}\label{eq:pczCalc}
	\ln p(\mathbf{z}|c)=\ln\frac{1}{\bm{\sigma}_c}
	-\frac{1}{2}\left[\ln(2\pi)
	+\frac{\bm{\mu}_c^2}{\bm{\sigma}_c^2}
	-\frac{2\bm{\mu}_c}{\bm{\sigma_c}^2}
	+\frac{\mathbf{z}^2}{\bm{\sigma}_c^2}
	\right],
\end{equation}
whose operators are all element-wise. If the model becomes extremely confident that a sample $\mathbf{x}$ (represented by latent embedding $\mathbf{z}$) does not belong to cluster $c$, then the value of $\ln p(\mathbf{z}|c)$ can be rather small, which potentially leads to numerical overflow when computing $e^{\ln p(\mathbf{z}|c)}$. Pretraining generally pushes $\ln p(\mathbf{z}|c)$ to either $0$ or $-\infty$, resulting in a much higher probability of having \code{NaN} loss values. GMVAE also suffers from this problem due to the formulation of $p(c|\mathbf{z},\mathbf{z}')$:
\begin{equation}\label{eq:NanLossGMVAE}
	p(c|\mathbf{z},\mathbf{z}')=\frac{p(c)e^{\ln p(\mathbf{z}|\mathbf{z}',c)}}{\sum_{c'=1}^C p(c')e^{\ln p(\mathbf{z}|\mathbf{z}',c')}}\ ,
\end{equation}
where $p(\mathbf{z}|\mathbf{z}',c)=\mathcal{N}(\mathbf{z}|\bm{\mu}_c(\mathbf{z}'),\bm{\sigma}_c^2(\mathbf{z}')\mathbf{I})$. M1+2 largely refrain from this problem by replacing the categorical distribution with a continuous one approximated by NN. The na\"ive fix of limiting the range of input values and initial model weights is suboptimal since it does not address the root cause of \code{NaN} loss.

Since each mini-batch training step uses $L$ samples, for unification, we use $\mathbf{V}\in\mathbb{R}_{c\times L}^-$ to denote the actual matrix of $\ln p(\mathbf{z}|c)$ and $\ln p(\mathbf{z}|\mathbf{z}',c)$ in VaDE and GMVAE, respectively. If we manage to eliminate overly small elements in $\mathbf{V}$, the \code{NaN} loss can also be prevented. One intuitive solution is clipping the values in $\mathbf{V}$, but it can easily halt the training due to the vanishing gradient. We instead propose an ``inverse min-max transform'' operator that re-scales each element in $\mathbf{V}$ to the range of $[-\lambda, 0]$:
\begin{equation}\label{eq:minmaxNorm}
	\widetilde{\mathbf{V}}=\frac{\lambda[\mathbf{V}-\min(\mathbf{V})]}{\max(\mathbf{V})-\min(\mathbf{V})}\ .
\end{equation}
The functions $\min(\cdot)$ and $\max(\cdot)$ return the minimum and maximum element from a matrix, respectively. $\lambda$ is a scaling factor, to which we suggest to assign a value between $20$ and $50$. With this, the properties of \textit{stability} and \textit{simplicity} of VDC training are attained by guaranteeing the convergence while entirely preventing the occurrence of \code{NaN} losses.

%-----------
\subsection{The Proposed Holistic Algorithm: S3VDC}
\label{holisticS3VDC}
\begin{algorithm}[tb]
\caption{S3VDC: Simple, Scalable, and Stable VDC}
\label{algo:s3vdc}
\textbf{Input}: dataset $\mathbf{X}$ and noisy dataset $\hat{\mathbf{X}}$ with $N$ samples \\
\textbf{Parameter}: $L$, $\gamma$, $T_\gamma$, $T_\beta$, $T_\text{s}$, $M$, $\lambda$, $C$, $k$, $f_{\bm{\theta}}(\cdot)$, $g_{\bm{\phi}}(\cdot)$ \\
\textbf{Output}: Trained $\bm{\theta}$, $\bm{\phi}$, $\bm{\pi}$, $\bm{\mu}_c$, $\bm{\sigma}_c^2$
\begin{algorithmic}[1] %[1] enables line numbers
\FOR{mini-batch step $t=1$ \TO $T_\gamma$} 
 \STATE Perform $\mathbf{\gamma}$-training using objective in Equation~\eqref{eq:gammaPretrain} w.r.t. Equation~\eqref{eq:ELBO_VaDE}: \\$\mathcal{L}_\gamma=\mathbb{E}_{q(\mathbf{z},c|\hat{\mathbf{x}})}[\ln p(\mathbf{x}|\mathbf{z})]-\gamma\mathcal{D}(q(\mathbf{z},c|\hat{\mathbf{x}})||p(\mathbf{z},c))$;
\ENDFOR
\STATE Initialize $\bm{\pi}$, $\bm{\mu}_c$, $\bm{\sigma}_c^2$ with GMM using $k\!\times\!L$ samples;
\FOR{$m=1$ \TO $M$} 
 \STATE $T_m=T_\gamma+(m-1)(T_\beta+T_\text{s})$
 \FOR{mini-batch $t=T_m+1$ \TO $T_m+T_\beta+T_\text{s}$} 
  \STATE Calculate $\ln p(\mathbf{z}|c)$ with Equation~\eqref{eq:pczCalc};
  \STATE Calculate $\widetilde{\ln}\ p(\mathbf{z}|c)$ with Equation~\eqref{eq:minmaxNorm}: $\widetilde{\ln}\ p(\mathbf{z}|c)=\frac{\lambda[\ln p(\mathbf{z}|c)-\min(\ln p(\mathbf{z}|c))]}{\max(\ln p(\mathbf{z}|c))-\min(\ln p(\mathbf{z}|c))}$;
  \STATE Calculate $q(c|\hat{\mathbf{x}})$ with Equation~\eqref{eq:qcxApprox}: $q(c|\hat{\mathbf{x}})= \frac{p(c)e^{\widetilde{\ln}\ p(\mathbf{z}|c)}}{\sum_{c'=1}^C p(c')e^{\widetilde{\ln}\ p(\mathbf{z}|c')}}$;
  \IF {$t\leq T_m+T_\beta$}
   \STATE Calculate $\beta=\gamma+[(t-T_m) / T_\beta]^3$;
   \STATE Perform $\beta$-annealing using objective in Equation~\eqref{eq:anneal} w.r.t. Equation~\eqref{eq:ELBO_VaDE}: \\ $\mathcal{L}_\beta=\mathbb{E}_{q(\mathbf{z},c|\hat{\mathbf{x}})}[\ln p(\mathbf{x}|\mathbf{z})]-\beta[\mathcal{D}(q(c|\hat{\mathbf{x}})||p(c))+\mathcal{D}(q(\mathbf{z}|\hat{\mathbf{x}})||p(\mathbf{z}|c))]$;
  \ELSE
   \STATE Optimize denoising Equation~\eqref{eq:ELBO_VaDE}:\\ $\mathcal{L}_{\text{VaDE}}=\mathbb{E}_{q(\mathbf{z},c|\hat{\mathbf{x}})}[\ln p(\mathbf{x}|\mathbf{z})]-\mathcal{D}(q(c|\hat{\mathbf{x}})||p(c))+\mathcal{D}(q(\mathbf{z}|\hat{\mathbf{x}})||p(\mathbf{z}|c))$;
  \ENDIF
 \ENDFOR
\ENDFOR
\STATE \textbf{return} $\bm{\theta}$, $\bm{\phi}$, $\bm{\pi}$, $\bm{\mu}_c$, $\bm{\sigma}_c^2$
\end{algorithmic}
\end{algorithm}

The solutions presented above (Section~\ref{sec:gammatraining}$-$\ref{sec:minmaxNorm}) are generic for VDCs. However, for concreteness, we choose to focus on VaDE due to its ability to finetune the cluster weights $\bm{\pi}$. We propose S3VDC (simple, scalable, and stable VDC) that incorporates all of the introduced solutions to a vanilla VaDE. Algorithm~\ref{algo:s3vdc} illustrates the details of S3VDC, where the lines $2$, $13$, and $15$ respectively represent $\mathbf{\gamma}$-training, $\beta$-annealing, and vanilla VaDE optimization. We empirically discover that applying the inverse min-max transform to line 13 and 15 is mandatory, while it is optional for $\mathbf{\gamma}$-training (line 2). 

Like all VDC models, S3VDC is optimized with stochastic mini-batch steps, thus there is no known formal guarantee on the number of steps required to achieve convergence. However, the time complexity of the $t$-th mini-batch optimization step (i.e. line 1 and 7 in Algorithm~\ref{algo:s3vdc}), $\mathcal{O}_t$, is dominated by the amount of matrix multiplications in the NN architecture. For a standard $o_l$-layered MLP with $o_n$ neurons in each layer, $\mathcal{O}_t$ is largely $\mathcal{O}(L\!\times\!o_l\!\times\!o_n^2)$. Besides, line $4$ in Algorithm~\ref{algo:s3vdc} is a special training step that finds optimal initial values for $\bm{\pi}$, $\bm{\mu}_c$, $\bm{\sigma}_c^2$ using GMM EM (expectation-maximization) algorithm\footnote{Specifically the implementation in https://github.com/scikit-learn}; and the computational complexity of each EM step is approximately $\mathcal{O}_{\text{GMM}}(k\!\times\!L\!\times\!C\!\times\!d_\mathbf{z}^3)$, where $d_\mathbf{z}$ is the dimension of $\mathbf{z}$. Since GMM initialization is only executed once, the overall time complexity of each training step is equivalent to $\mathcal{O}_t$; but $\mathcal{O}_{\text{GMM}}$ may become a critical training bottleneck when a large number of samples (e.g. $k\!\times\!L>1\!\times\!10^6$) are used for GMM initialization. We will investigate the optimal $k\!\times\!L$ in Section~\ref{exp:scalability}. Despite the implementation differences in various software libraries, the space complexity of S3VDC primarily consists of three parts: mini-batch samples ($\mathcal{O}(L\!\times\!D)$), DNN (such as weights, gradients, and intermediate cache that are often consistent among VDCs as long as they share the same network architecture), and GMM initialization ($\mathcal{O}(k\!\times\!L\!\times\!C\!\times\!d_\mathbf{z}^2)$~\cite{zhou2018gaussian}).

\begin{table}[tb!]
\footnotesize\addtolength{\tabcolsep}{1.2pt}\renewcommand{\arraystretch}{1}
\centering
\caption{Dataset specifications (top) and S3VDC hyper-parameters (bottom).} 
\label{tab:exp_setting}
\begin{threeparttable}
\begin{tabular}{c|c|c|c|c}
\bottomrule

Specification & InertialHAR \cite{anguita2013public} & MNIST \cite{lecun1998gradient} & Fashion \cite{xiao2017fashion} & King10M
\\
\hline
\# sample & 10,299 & 70,000 & 70,000 & 9,666,892
\\
\# feature\tnote{*} & 9$\times$128 & 28$\times$28 & 28$\times$28 & 8$\times$30
\\
\# cluster: $C$ & 6 & 10 & 10 & N/A (6)\tnote{$\dagger$}
\\
\hline
dimension of $\mathbf{z}$: $d_\mathbf{z}$ & 4 & 8 & 6 & 10
\\
mini-batch size: $L$ & 1,024 & 128 & 64 & 1,024 
\\
GMM initialization steps: $k$ & 5 & 200 & 400 & 750
\\
initial learning rate\tnote{$\ddagger$} & 1.5$\times$10$^{\text{-}3}$ & 2$\times$10$^{\text{-}3}$ & 2$\times$10$^{\text{-}3}$ & 1.5$\times$10$^{\text{-}3}$
\\
$\gamma$-training weight: $\gamma$ & 5$\times$10$^{\text{-}6}$ & 5$\times$10$^{\text{-}4}$ & 5$\times$10$^{\text{-}3}$ & 5$\times$10$^{\text{-}4}$
\\
$\gamma$-training steps: $T_\gamma$ & 6$\times$10$^{3}$ &  1$\times$10$^{5}$ & 1$\times$10$^{5}$ & 2.5$\times$10$^{4}$
\\
$\beta$-annealing steps: $T_\beta$ & 2.5$\times$10$^{3}$ & 9$\times$10$^{3}$ & 9$\times$10$^{3}$ & 4.5$\times$10$^{3}$ 
\\
static steps: $T_\text{s}$ & 5$\times$10$^{2}$ & 1$\times$10$^{3}$ & 1$\times$10$^{3}$ & 5$\times$10$^{2}$ 
\\
\# period: $m$ & 2 & 10 & 10 & 6
\\
\toprule
\end{tabular}
\begin{tablenotes}
        \scriptsize
        \item[*] Each sample is transformed into a shape of $L\!\times\!28\!\times\!28\times$(\#channel) via zero padding and resizing (interpolation). InertialHAR and King10M treat each timeseries as a channel.
        \item[$\dagger$] King10M has no ground-truth label, thus we set $C\!=\!6$ based on the results in Section~\ref{nCluster} (Table~\ref{tab:nCluster}). 
        \item[$\ddagger$] Exponential learning rate decay with a terminating value of $1\times10^{-6}$.
        
        \vspace{-5pt}
\end{tablenotes}
\end{threeparttable}
\end{table}

%% === SEC: EXPERIMENTS & EXPLORATIONS ==============
\section{Experiments and Explorations}
\label{sec:exp}
In Section~\ref{sec:gammatraining} and \ref{sec:minmaxNorm}, we discussed \textit{simplicity} (P6). This section intends to evaluate \textit{scalability} (P7), \textit{stability} (P8), and other industrial properties (P1-P5) of S3VDC using four unstructured datasets (Table~\ref{tab:exp_setting}). We zero-center the original inertial signals in InertialHAR~\cite{anguita2013public} dataset, but use MNIST~\cite{lecun1998gradient} and Fashion~\cite{xiao2017fashion} as they are. King10M dataset contains the daily behaviour of almost ten million game players that play Candy Crush Soda Saga (CCSS) \footnote{https://king.com/game/candycrushsoda}. Each sample contains counters (aggregated per day over a period of 30 days commencing from a randomly picked date) for eight types of in-game actions (e.g. win a level, buy an item, send a message, etc.) from an individual player. The individual players in King10M dataset are randomly sampled from the entire population of CCSS game players.

All experiments are carried out on the Google cloud platform with the hardware scale tier \code{BASIC-GPU}\footnote{https://cloud.google.com/ml-engine/docs/machine-types}. The noise added to input follows a Gaussian distribution with a standard deviation (STD) of $5\times10^{-9}$. We determine the hyper-parameters (lower part of Table~\ref{tab:exp_setting}) using a discrete random search with early stopping, which is carried out using an in-house developed ML platform\footnote{S3VDC is \href{https://github.com/king/s3vdc}{open-sourced} in a self-contained manner that is independent of any ML platforms, cloud computing services, or data-feed paradigms.}. During that process, $\lambda$ in Equation~\eqref{eq:minmaxNorm} is uniformly set to 50, and GMM initialization takes $1\times10^4$ EM iteration steps; we also discovered that setting $u$=3 in Equation~\eqref{eq:anneal} generally works well on all of the datasets presented in Table~\ref{tab:exp_setting}. The CNN architecture is the same as~\cite{guo2017deep}. We used the same setting as S3VDC for VaDE and DCEC when applicable for a fair comparison. All reported scores are averaged over five trials with randomly initialized model weights.

\begin{table}[tb!]
\footnotesize\addtolength{\tabcolsep}{5pt}\renewcommand{\arraystretch}{1.2}
\centering
\caption{Comparison of clustering accuracy and stability over benchmark datasets.} 
\label{tab:stability}
\begin{threeparttable}
\begin{tabular}{c|c|c|c|c|c|c}
\bottomrule
\multirow{2}{*}{Dataset}
& 
\multicolumn{2}{c|}{DCEC\tnote{*}}
& 
\multicolumn{2}{c|}{VaDE\tnote{*}}
& 
\multicolumn{2}{c}{S3VDC}
\\
\cline{2-7}
& Accuracy & STD & Accuracy & STD & Accuracy & STD 
\\
\hline
InertialHAR 
& 0.5594 & 0.0682 & 0.5900 & 0.0590 & {\bf0.6705} & {\bf0.0130}
\\
MNIST
& 0.8690 & 0.0890 & 0.8400 & 0.1060 & {\bf0.9360} & {\bf0.0181}
\\
Fashion & 0.5180 & 0.0380 & 0.5500 & 0.0420 & {\bf0.6053} & {\bf0.0091} 
\\
\toprule
\end{tabular}
\begin{tablenotes}
        \scriptsize
        \item[*] For DCEC, we initialize $k$-means using the same number of samples as S3VDC. For VaDE, we do not use any pre-trained model and discard the runs with \code{NaN} losses.
        \vspace{-5pt}
\end{tablenotes}
\end{threeparttable}
\end{table}

%----------
\subsection{Clustering Stability and Accuracy}
\label{clusteringStability}
The stability is measured by STD of accuracy, which is calculated by finding the best mapping between cluster assignments and labels. Table~\ref{tab:stability} illustrates that S3VDC not only reaches the state-of-the-art clustering accuracy but also obtains a substantial decrease in STD. This guarantees extremely stable clustering results while maintaining high accuracy with respect to the target labels.

\begin{figure}[tb!]
\centering
\scalebox{0.7}{\includegraphics{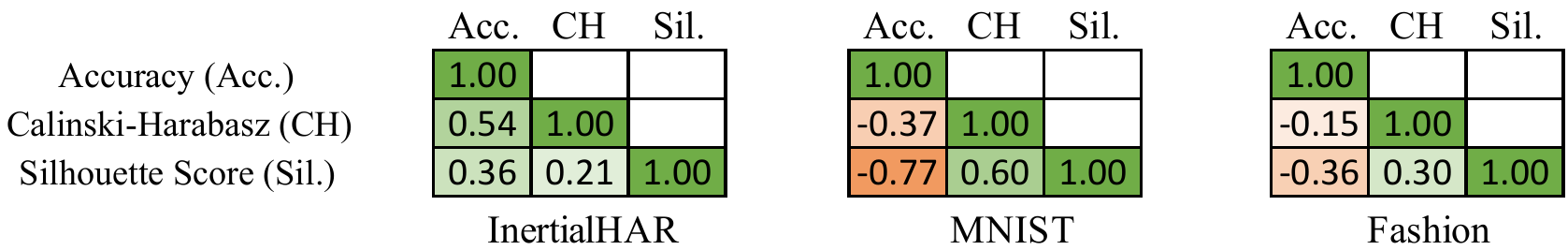}}
\caption{Pair-wise correlation of clustering metrics for S3VDC.}
\label{fig:metricCorr}
\end{figure}

\begin{figure}[tb!]
\centering
\scalebox{0.4}{\includegraphics{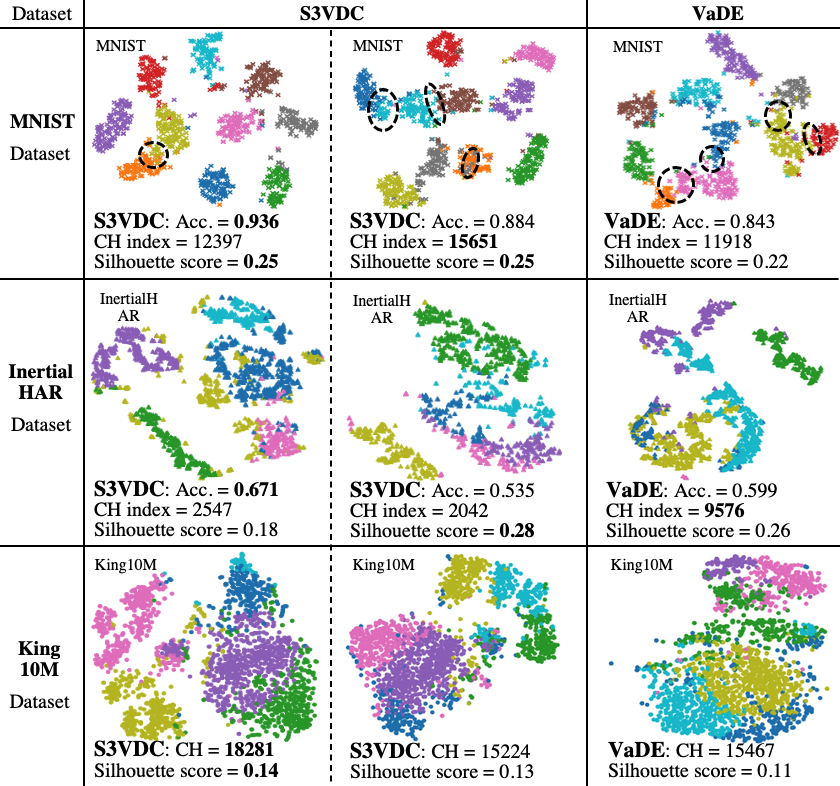}}
\caption{The t-SNE visualization of the learned $\mathbf{z}$ colored with predicted clusters. To enable VaDE training, we apply the same mini-batch GMM initialization as S3VDC.}
\label{fig:latentVis}
\end{figure}

%------------
\subsection{Unsupervised Model Selection}
\label{modelSelection}
Unlike the labeled benchmark datasets, it is much harder to evaluate clustering results without any ground truth label. To that end, we investigated the feasibility of model selection using unsupervised metrics and latent disentanglement. We considered all successful trials from hyper-parameter search.

{\bf Correlation among clustering metrics}:
Figure~\ref{fig:metricCorr} illustrates how much supervised metric (accuracy) and unsupervised metrics (Calinski-Harabasz index and Silhouette score) agree by calculating the pair-wise Spearman rank correlation for the selected metrics. Calinski-Harabasz (CH) index is defined as the ratio between within-cluster dispersion and between-cluster dispersion, which is higher when clusters are dense and well-separated. Silhouette score measures how similar a sample is to its own cluster compared to other clusters. This score ranges from $-1$ to $1$, where higher values indicate better clustering quality. We observe that supervised and unsupervised metrics do not correlate in the same way on different datasets, yet unsupervised metrics almost always agree. So, in addition to unsupervised clustering metrics, examining the latent disentanglement becomes inevitable for unsupervised model selection.

{\bf Disentanglement of latent embedding}:
\label{latentVis}
In Figure~\ref{fig:latentVis}, we demonstrate that the disentanglement degree (visualized together with the predicted clusters) of the learned latent embedding $\mathbf{z}$ is an indicator of clustering quality. The left column contains our selected S3VDC models; and the S3VDC models in middle column are obtained during the hyper-parameter search and considered sub-optimal. The right-most column of Figure~\ref{fig:latentVis} represents the best VaDE models obtained on each dataset. It can be seen that disentanglement degree sometimes leads to selecting a different model than what unsupervised metrics might suggest; for an instance, we would never select the left-most S3VDC model for InertialHAR dataset by only look at either CH index or Silhouette score. Thus, we used a combination of unsupervised clustering metrics and latent disentanglement to choose the best hyper-parameters for S3VDC on King10M dataset.

\begin{figure}[tb!]
\centering
\scalebox{0.66}{\includegraphics{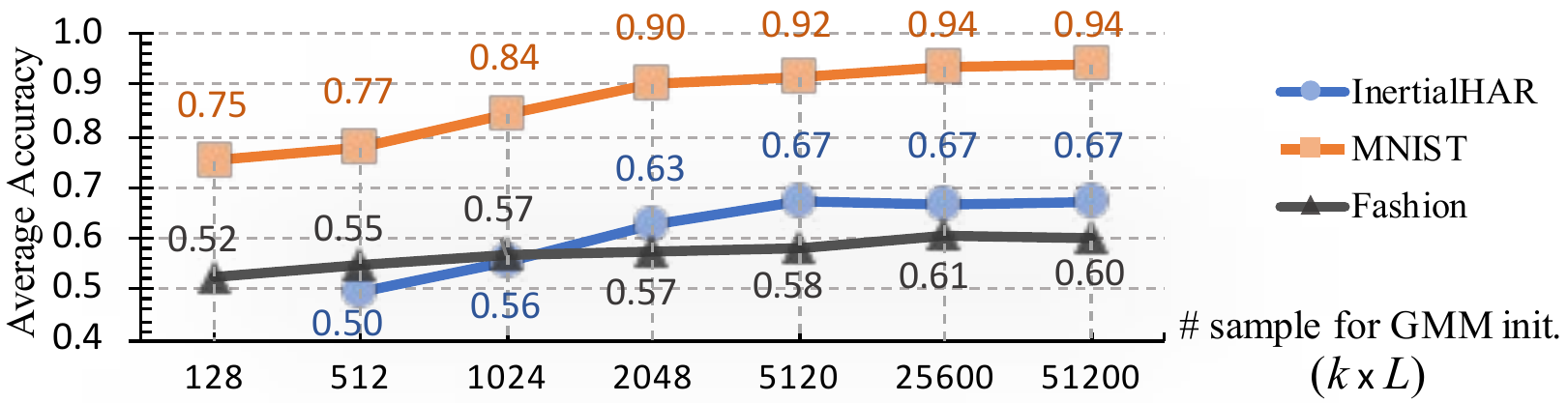}}
\caption{S3VDC accuracy in relation to the number of samples for GMM initialization.}
\label{fig:scalability}
\end{figure}

%-----------
\subsection{Training Scalability}
\label{exp:scalability}
In Section~\ref{proposalScale}, we proposed to use only a Monte Carlo subset (of size $k\!\times\!L$) of the entire dataset to initialize the global GMM parameters. Figure~\ref{fig:scalability} shows that the clustering accuracy of S3VDC monotonically increases with the value of $k\!\times\!L$, and saturates at different places. Observed from Table~\ref{tab:scalability}, it is generally sufficient to use less than half of the dataset for GMM initialization, which leads to improved time complexity and training scalability. Specifically, King10M merely needs about 8\% of all samples to reach the saturated Silhouette and CH scores, while VaDE keeps failing at the full-scale GMM initialization step.

\begin{table}[tb!]
\footnotesize\addtolength{\tabcolsep}{4.5pt}\renewcommand{\arraystretch}{1.2}
\centering
\caption{The average time consumption (seconds) of GMM initialization.} 
\label{tab:scalability}
\begin{threeparttable}
\begin{tabular}{c|c|c|c|c}
\bottomrule
VDC Models & InertialHAR & MNIST & Fashion & King10M
\\
\hline
VaDE & 0.66 s & 2.44 s & 2.42 s & Failed to initialize
\\
S3VDC\tnote{*} & 0.34 s (50\%) & 0.98 s (37\%) & 1.04 s (37\%) & 9.15 s (8\%)
\\
\toprule
\end{tabular}
\begin{tablenotes}
        \scriptsize
        \item[*] The the minimum percentage of dataset required to achieve the accuracy plateau is indicated in parentheses. VaDE uses the entire dataset for GMM initialization.
        \vspace{-5pt}
\end{tablenotes}
\end{threeparttable}
\end{table}

%-----------
\subsection{Optimal Number of Clusters}
\label{nCluster}
To determine the optimal number of clusters, $C$, without using any label information, we report the Monte Carlo estimated marginal likelihood $-\ln p(\mathbf{x})$ for S3VDC over the test sets ($\sim$10\% of each dataset) for different values of $C$. From Table~\ref{tab:nCluster}, we find that $-\ln p(\mathbf{x})$ often reaches its minimum around the ``preferred'' $C$ on benchmark datasets. The true number of clusters can become less obvious if hard-to-disentangle clusters exist (e.g., Fashion dataset visualized in Figure~\ref{fig:latentVis}).

\begin{table}[tb!]
\footnotesize\addtolength{\tabcolsep}{2.5pt}\renewcommand{\arraystretch}{1.1}
\centering
\caption{The S3VDC marginal likelihood $-\ln p(\mathbf{x})$ vs. the target cluster number $C$.}
\label{tab:nCluster}
\begin{tabular}{c|c|c|c|c|c|c|c|c|c}
\bottomrule
Dataset & C=4 & C=5 & C=6 & C=7 & C=8 & C=9 & C=10 & C=11 & C=12
\\
\hline
InertialHAR & 23.89 & 23.81 & {\bf22.97} & 23.54 & 24.05 & - & - & - & -
\\
MNIST & - & - & - & - & 89.61 & 89.09 & {\bf89.08} & 89.09 & 89.32
\\
Fashion & - & - & - & 58.46 & 58.35 & {\bf57.97} & 57.98 & 58.01 & 58.17
\\
King10M & 39.77 & 39.94 & {\bf39.68} & 39.95 & 40.11 & - & - & - & -
\\
\toprule
\end{tabular}
\end{table}

%-----------
\subsection{Model Interpretation without Label}
\label{modelInterpretation}
\begin{figure}
\centering
\scalebox{0.44}{\includegraphics{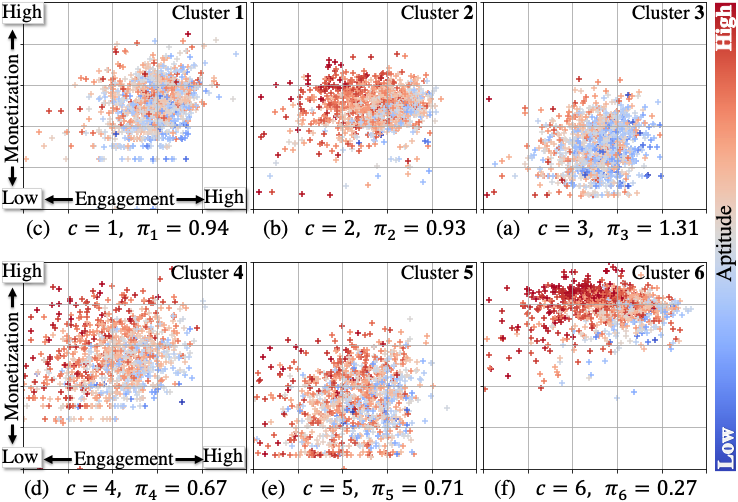}}
\caption{Interpret the generated player data from a S3VDC model (trained on King10M dataset) towards three user KPIs: engagement, revenue, and aptitude.}
\label{fig:interpCluster}
\end{figure}

When evaluating the performance of S3VDC over large unlabeled datasets like King10M, we suggest to simultaneously look at 1) unsupervised metrics, 2) the joint visualization of latent disentanglement and cluster prediction, and 3) the marginal likelihood. Accordingly, we selected the best S3VDC model (see bottom-left in Figure~\ref{fig:latentVis}) trained on King10M. To assign meaningful labels to all six clusters for King10M, we first generate 1,000 users by sampling the latent embedding from each cluster using the learned S3VDC model, then we plot them in a three-dimensional KPI (key performance indicator) space for each cluster. As shown in Figure~\ref{fig:interpCluster}, the three KPIs describe 1) engagement (horizontal axis): the level of commitment to the gaming activities; 2) monetization (vertical axis): the amount of in-game spending; and 3) aptitude (anchor color): the probability of wining game levels. The KPI calculation (from the generated time-series) and the actual values are considered to be sensitive proprietary data and therefore removed from the plot. We observe that clusters 1, 2, and 6 contain slightly more active users; clusters 2 and 6 represent mostly users with high aptitude, while clusters 1 and 3 have mainly low-aptitude users; the ranking of in-game spending is 6$>$1,2,4$>$3,5. Moreover, cluster 3 and 6 respectively have the highest and lowest cluster weights.

%% === SEC: CONCLUSION ==============
\section{Conclusion}
\label{sec:conclusion}
In this paper, we challenge the state-of-the-art in DC solutions in their applicability in industry. To that end, we argue for eight properties that we believe DC methods need to hold to be able to fill in the gap of applicability. To address this gap, we propose, S3VDC, a simple stable and scalable VDC algorithm. Our experiments show that S3VDC outperforms the state-of-the-art VDC in terms of stability on standard datasets. In addition, we show that the proposed method is scalable since it requires half of samples needed for initialization compared to the state-of-the-art VDC. Finally, we experimentally demonstrate how S3VDC facilitates unsupervised model selection and interpretation.

Future work includes sensitivity analysis of hyper-parameters in S3VDC and utilizing generative processes for quantitative evaluation.

% ---- Bibliography ----
%
% BibTeX users should specify bibliography style 'splncs04'.
% References will then be sorted and formatted in the correct style.
\bibliographystyle{splncs04}
\bibliography{ref}
\end{document}